\documentclass[10pt,twocolumn,letterpaper]{article}

\usepackage{cvpr}              
\usepackage{multirow} 
\usepackage{pifont}
\definecolor{cvprblue}{rgb}{0.21,0.49,0.74}
\usepackage[pagebackref,breaklinks,colorlinks,allcolors=cvprblue]{hyperref}
\def\paperID{*****} 
\def\confName{CVPR}
\def\confYear{2025}

\title{From Pixels to Graphs: using Scene and Knowledge Graphs \\ for HD-EPIC VQA Challenge}

\author{Agnese Taluzzi\textsuperscript{1}\thanks{Equal Contribution.}
\quad
Davide Gesualdi\textsuperscript{1}$^*$
 \quad
Riccardo Santambrogio\textsuperscript{1}
\quad
Chiara Plizzari\textsuperscript{1}\\
\vspace*{-8pt} 
\quad
Francesca Palermo\textsuperscript{2}
\quad
Simone Mentasti\textsuperscript{1}
\quad
Matteo Matteucci\textsuperscript{1}
\quad
\vspace*{8pt} 
\and
\textsuperscript{1}Politecnico di Milano
\textsuperscript{2}EssilorLuxottica
}

\begin{document}
\maketitle
\begin{abstract}
This report presents SceneNet and KnowledgeNet, our approaches developed for the HD-EPIC VQA Challenge 2025. SceneNet leverages scene graphs generated with a multi-modal large language model (MLLM) to capture fine-grained object interactions, spatial relationships, and temporally grounded events. In parallel, KnowledgeNet incorporates ConceptNet’s external commonsense knowledge to introduce high-level semantic connections between entities, enabling reasoning beyond directly observable visual evidence. 
Each method demonstrates distinct strengths across the seven categories of the HD-EPIC benchmark, and their combination within our framework results in an overall accuracy of 44.21\% on the challenge, highlighting its effectiveness for complex egocentric VQA tasks.

\end{abstract}    
\section{Introduction}
\label{sec:intro}


\begin{figure}
    \centering
    \includegraphics[width=\linewidth]{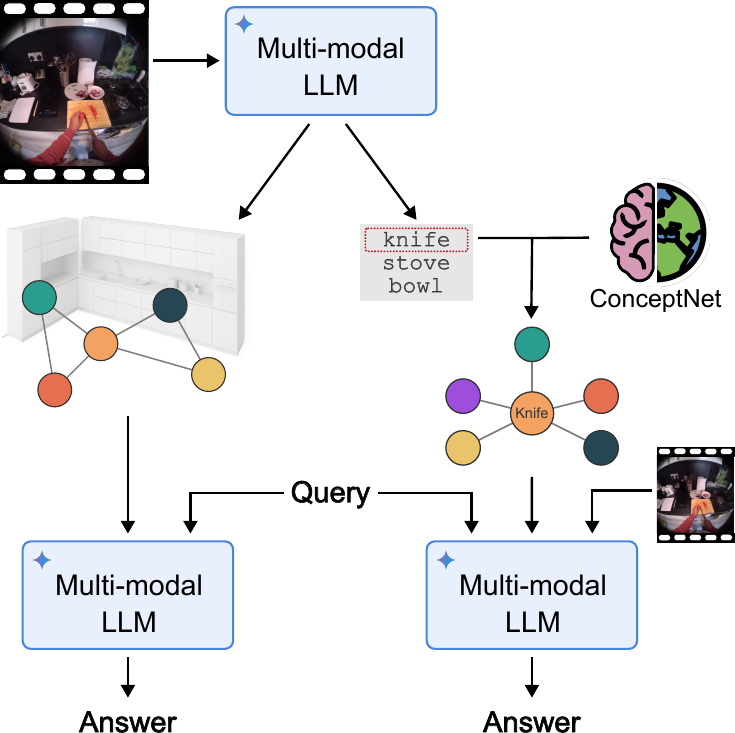}
    \caption{\textbf{Overview of our VQA method.} Given an input video, we use a multi-modal large language model (LLM) to construct two structured representations: (1) a scene graph capturing spatial relationships between objects (SceneNet), and (2) a knowledge graph for each object based on ConceptNet \cite{speer2018conceptnet55openmultilingual} (KnowledgeNet). These structured representations, together with a natural language question, are fed into the multi-modal LLM to generate an answer.}
    \label{fig:overview}
    \vspace{-20pt}
\end{figure}

Egocentric videos present unique challenges due to rapid camera motion from head and body movements and a limited, shifting field of view. These factors lead to frequent occlusions and partial observations, complicating reliable reasoning about interactions and temporal dynamics, especially when critical context lies outside the visible frame~\cite{grauman2022ego4d, Damen2021EPIC}. Visual question answering (VQA) on egocentric video suffers from these difficulties. First, reasoning must be performed over long sequences of frames, resulting in large amounts of data and tokens that must be efficiently processed. Second, most state-of-the-art multi-modal large language models (MLLMs), such as Gemini~\cite{du2023gemini}, are pretrained on extensive datasets that primarily consist of third-person or non-egocentric content, limiting their ability to generalize to the unique characteristics of egocentric video~\cite{alayrac2022flamingo, li2022blip}. These present a significant obstacle for current models, which often rely on frame-level representations and lack effective mechanisms to incorporate structured semantic knowledge for deeper reasoning about object affordances, state changes, and causal dynamics.

To address these challenges, we turn to neuro-symbolic AI, a composite framework that seeks to merge neural network-based methods with symbolic knowledge-based approaches~\cite{N_d’Avila_2023}. It is founded on the premise of attaining the complementary benefits of both approaches, integrating the neural network capabilities of direct training from raw data and robustness against faults in the underlying data with the symbolic reasoning capabilities that enable them to reason about abstract concepts, extrapolate from limited data and generate explainable results~\cite{yi2018neurosymbolic, mao2019neuro, hudson2019learning, Choi_2024}. 
This makes them especially suitable for challenging domains like egocentric video analysis, where semantic understanding, commonsense reasoning, and generalization to the egocentric domain are essential.

In this work, we propose a framework that addresses key challenges in egocentric VQA by incorporating two complementary neuro-symbolic abstractions: \textit{scene graphs} and \textit{commonsense knowledge graphs}, instantiated through dedicated modules we call \textbf{SceneNet} and \textbf{KnowledgeNet}, respectively. Scene graphs, originally introduced for static image understanding~\cite{johnson2015image} and extended to dynamic video domains~\cite{xu2017scenegraph, rodin2023actionscenegraphslongform, peddi2025unbiasedrobustspatiotemporalscene}, encode objects, attributes, spatial relationships, and interactions within frames in a structured symbolic format, enabling reasoning over abstracted scene representations. These capabilities are leveraged by SceneNet to capture fine-grained spatial and relational information. Commonsense knowledge graphs such as \textit{ConceptNet}~\cite{speer2018conceptnet55openmultilingual} provide external semantic knowledge about typical object interactions, affordances, and causal events~\cite{sap2019atomic, bosselut2019comet, ilharco2022patching, Mohammadi_Hong_Qi_Wu_Pan_Shi_2024, kundu2024algoobjectgroundedvisualcommonsense}, which KnowledgeNet integrates to support inference beyond direct visual evidence. An overview of the approach is provided in Figure \ref{fig:overview}. 


We detail the data processing pipeline, scene graph extraction methods, graph-based reasoning strategies, and provide experimental results across various configurations.

\section{Our Approach}
\label{sec:method}

\begin{figure}
    \centering
    \includegraphics[width=\linewidth]{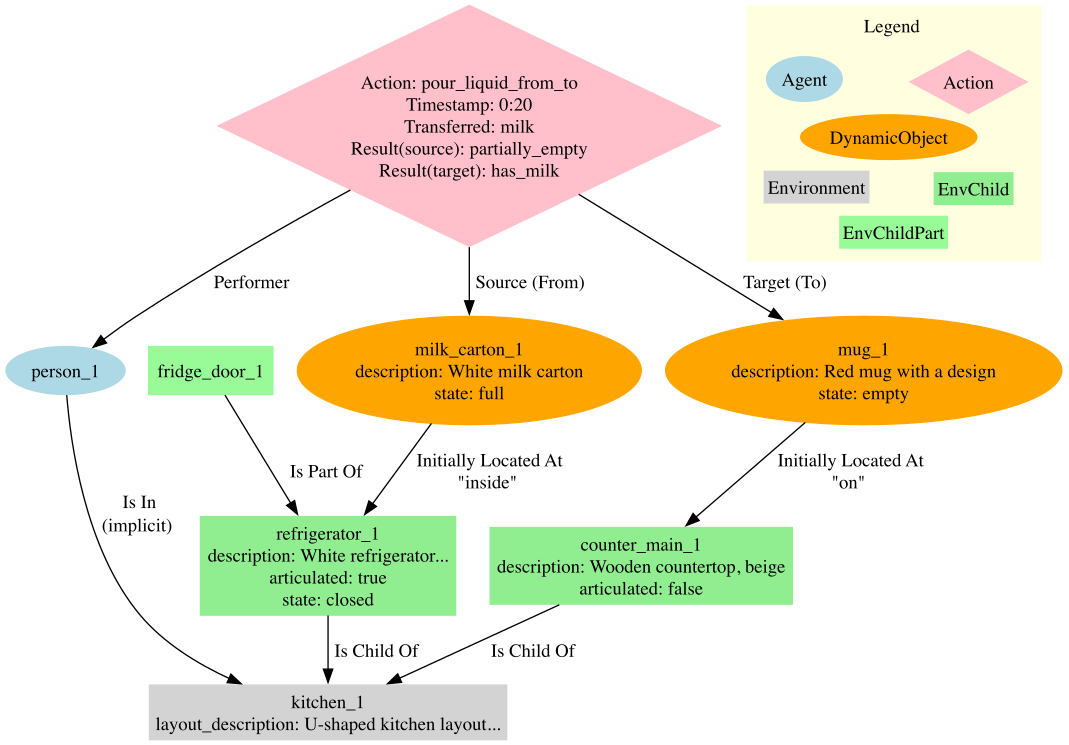}
    \caption{\textbf{SceneNet.} Example SceneNet graph illustrating an initial state with various entities (Agent, Environment, Dynamic Objects) and their relationships, followed by a  ``pour milk" action.}
    \label{fig:scene_graph_example}
    \vspace{-10pt}
\end{figure}
Our approach uses an MLLM conditioned on question, video (optional), and graph-based inputs: either SceneNet (visual-spatial relations) or KnowledgeNet (semantic associations from ConceptNet).
While both graphs are rooted in MLLM-extracted objects from the video or question, they differ in structure and purpose: SceneNet is visually grounded, whereas KnowledgeNet expands from a root object using ConceptNet~\cite{speer2017conceptnet} to capture implicit, external knowledge semantically related to the object.

The two graph construction pipelines are illustrated in Section \ref{sec:scenenet} and Section \ref{sec:knowledgenet} respectively.

\subsection{SceneNet}
\label{sec:scenenet}
\subsubsection{Graph Generation}
SceneNet models video segments as structured \textit{scene graphs} (objects, attributes, spatial/temporal relations).

We extract scene graphs from egocentric videos using prompted generation with an MLLM guiding the model to produce structured JSON outputs. According to this schema, a target scene graph $\mathcal{G}$ is to be structured as a tuple:
\begin{equation}
\mathcal{G} = (N, E_B, A)
\end{equation}
where $N$ is the set of nodes (entities), $E_B$ is a set of binary edges representing direct, structural relationships, and $A$ is a set of action relationships (hyperedges) representing interactions derived from the event timeline.

The conceptual design of the entities $N$ and their structural relationships $E_B$ draws inspiration from detailed 3D scene representations~\cite{armeni20193d}, adapted here for 2D video. The modeling of interactions $A$ as hyperedges is inspired by approaches utilizing situation hyper-graphs to capture multi-entity events for video understanding~\cite{urooj2023learning}.

The following paragraphs describe the expected constitution of $N$, $E_B$, and $A$ as per the provided schema.

\textbf{Core Entities (Nodes, $N$).}
Nodes $n \in N$ include: \texttt{Agent} (human actor), \texttt{Environment} (the kitchen), \texttt{EnvironmentChild} (fixed structural elements or large appliances, e.g., countertops, refrigerator), \texttt{EnvironmentChildPart} (movable sub-components like fridge doors), and \texttt{DynamicObjects} (movable or manipulated items, e.g. mugs, food items). Each has type-specific attributes (e.g. `description', `initial\_state', `articulated'--true if the object has movable parts).


\textbf{Direct Relationships (Binary Edges, $E_B$).}
Binary edges $e_b \in E_B$ define hierarchical (\texttt{Contains/Is Child Of}, \texttt{Has Part/Is Part Of}) and spatial (\texttt{InitiallyLocatedAt}) links. \texttt{CreatedFrom} edges track when a dynamic object is originated from another one (e.g., when an object is sliced into smaller parts)

\textbf{Interactions (Action Relationships / Hyperedges, $A$).}
Action hyperedges $a \in A$ represent multi-faceted, timestamped interactions, primarily defined by an \texttt{action} (e.g., ``take\_object", ``pour\_from\_to", ``stir"). Each action connects an \texttt{agent} (the performer) to various optional participant entities such as \texttt{source}, \texttt{target}, \texttt{location}, \texttt{tool}, or any newly created dynamic objects whose involvement can vary depending on the specific action. Key descriptive properties of the interaction, like the `resulting\_state\_of\_source' and `resulting\_state\_of\_target', are also captured within the hyperedge.



Figure~\ref{fig:scene_graph_example} provides a visual example of the scene graph representation used in SceneNet.




\subsection{KnowledgeNet}
\label{sec:knowledgenet}
\begin{figure}[t]
    \centering
    \includegraphics[width=\linewidth]{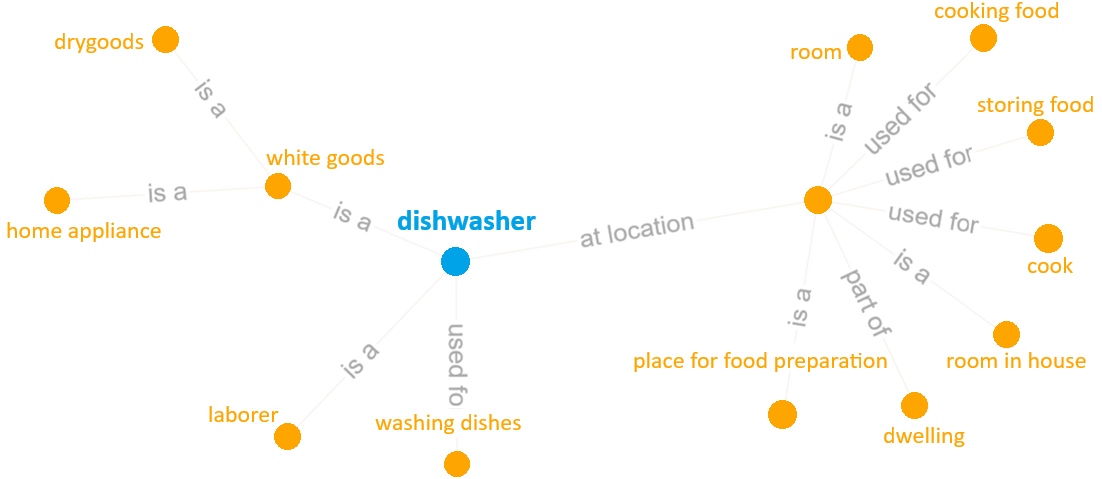}
    \caption{\textbf{KnowledgeNet. }Simplified knowledge graph rooted at the concept node \textit{dishwasher}, prior to domain filtering.}
    \label{fig:graph_conceptnet}
\end{figure}
{

To facilitate deeper reasoning about objects affordances, we construct symbolic \textit{commonsense knowledge graphs} rooted in objects identified in video clips, using ConceptNet as our external knowledge source.
ConceptNet provides concept nodes, each representing a word or short phrase in natural language, typically a common noun or verb in its undisambiguated form, linked through directed, labeled edges, also referred to as \textit{assertions}.
These nodes and assertions are used to expand object-centric graphs. We outline our graph generation pipeline below.

\textbf{Object Recognition and Root Node Selection.}
Objects are first detected using zero-shot object recognition with an MLLM, prompted with a dedicated template. Let $\mathcal{O} = \{o_1, \dots, o_m\}$ be the set of all detected object mentions in a scene. Each object $o_i$ serves as the root node for a knowledge graph $\mathcal{G}_i$.

\textbf{Knowledge Graph Definition.}
Each object-centric knowledge graph $\mathcal{G}_i$ is represented as a labeled, directed, and attributed graph defined as:
\begin{equation}
\mathcal{G}_i = (V_i, R, E_i),
\end{equation}
where $V_i$ is the set of concept nodes (e.g., ${\texttt{kitchen},\ \texttt{cupboard},\ \dots}$), $R$ is a predefined set of semantic relations (e.g., \texttt{used for}, \texttt{part of}, \texttt{has property}), and $E_i \subseteq V_i \times R \times V_i$ is the set of directed, labeled edges, referred to as assertions.

Each assertion $e \in E_i$ is a relational triple of the form:
\begin{equation}
e = (\mathrm{a}, \mathrm{r}, \mathrm{b}),
\end{equation}
where $\mathrm{a}, \mathrm{b} \in V_i$, $\mathrm{a} \ne \mathrm{b}$, and $\mathrm{r} \in R$ denotes the semantic relation linking the two nodes (e.g., \texttt{cupboard} \texttt{is used for} \texttt{storing dishes}).

Each graph $\mathcal{G}i$ is a subgraph of the global ConceptNet knowledge graph. 
That is, $V_i \subseteq V_{\text{CN}}$, $R \subseteq R_{\text{CN}}$, and $E_i \subseteq E_{\text{CN}}$, where $V_{\text{CN}}, R_{\text{CN}}, E_{\text{CN}}$ represent the full set of nodes, relations, and assertions in ConceptNet, respectively.

\textbf{Graph Construction.}
Graphs are constructed through a breadth-first expansion from the root node $o_i$ up to depth $d = 3$. 
We initialize the node and edge sets as:
\begin{equation}
    V_i^{(0)} = \{o_i\}, \quad E_i^{(0)} = \emptyset
\end{equation}

At each step $k \in \{0, \dots, d-1\}$, we expand the edge set using all ConceptNet assertions involving nodes from the current layer and valid relations:
\begin{equation}
    \tilde{E}_i^{(k+1)} = E_i^{(k)} \cup \left\{(v, r, v') \in E_{\text{CN}} \,\middle|\, v \in V_i^{(k)},\ r \in R \right\}
\end{equation}
To prevent redundancy, symmetric relations (i.e., \texttt{similar to} and \texttt{synonym}) are ignored for $k > 0$.

Each assertion $e$ in ConceptNet is associated with a confidence score, which we normalize to the interval $[0, 1]$ using a QuantileTransformer~\cite{scikit-learn}:
\begin{equation}
w: E_{CN} \rightarrow [0, 1]
\end{equation}
We retain only edges with $w(e) > 0.7$. When multiple edges connect the same node pair $(v, v')$, only the one with the highest weight is kept:
\begin{equation}
E_i^{(k+1)} = \bigcup_{(v, v') \in \mathcal{P}} \left\{ \arg\max_{e \in \{ (v, r, v') \in \tilde{E}_i^{(k+1)} \}} w(e) \right\}
\end{equation}
where 
\begin{equation}
    \mathcal{P} = \left\{ (v, v') \mid \exists r: (v, r, v') \in \tilde{E}_i^{(k+1)},\ w(e) > 0.7 \right\}
\end{equation}

Finally, we update the node set:
\begin{equation}
V_i^{(k+1)} = V_i^{(k)} \cup \{v' \mid (v, r, v') \in E_i^{(k+1)}\}
\end{equation}

\textbf{Path Extraction.}
Graphs are serialized into semantic paths, each formally defined as a sequence of the form:
\begin{equation}
P = \{v_0, \mathrm{r}_1, v_1, \mathrm{r}_2, \dots, \mathrm{r}_k, v_k\},
\end{equation}
starting from the root $\mathcal{V}_i^{(0)} = \{\mathrm{v}_0\} \subseteq \mathcal{O}$ and traversing through labeled relations. These are rendered into natural language templates (e.g., \textit{``cupboard is used for storing dishes''}) for inclusion in the MLLM prompt. 

\textbf{Domain Filtering.}
To mitigate lexical ambiguity and out-of-context expansion, we filter the textual paths by measuring their semantic relevance to a kitchen context. Let $S = \{s_1, \dots, s_n\}$ be a set of reference sentences describing kitchen-relevant affordances. Each path $P$ is embedded as $\mathbf{p} = \text{Enc}(P)$ using a Sentence Transformer model~\cite{reimers-2019-sentence-bert}\footnote{Specifically \texttt{NovaSearch/jasper\_en\_vision\_language\_v1}~\cite{zhang2025jasperstelladistillationsota}.}. Relevance is computed by:
\begin{equation}
\text{sim}(P) = \frac{1}{\left| S \right|} \sum_{s \in S} \cos(\mathbf{p}, \text{Enc}(s))
\end{equation}
where $\cos(\cdot, \cdot)$ denotes the cosine similarity between the path embedding $\mathbf{p}$ and the encoded sentence $\text{Enc}(s)$. 
Paths are ranked by cosine similarity and up to 30 top-ranked paths are retained for each object (or all paths if fewer than 30 exist), ensuring contextual coherence.

\subsection{Inference}

We perform inference by prompting the MLLM with the question, the encoded graph (SceneNet or KnowledgeNet), and optionally video (see Section \ref{sec:experiments} for details), to select one multiple-choice answer.
\section{Experiments}
\label{sec:experiments}
In this section, we explain the details for graph generation and the different input configurations used. We then report the results in Section \ref{sec:results}.
\paragraph{Implementation Details. }
All experiments use Gemini 2.0 Flash~\cite{GeminiFlash} 
for both graph generation and inference. For {SceneNet graph generation}, videos were sampled at 1 FPS and 480p resolution, then split into segments of up to 400 seconds. A scene graph (see Section~\ref{sec:scenenet}) was generated independently for each segment. To ensure temporal consistency, timestamps were adjusted globally across segments (e.g., “0:10” in a segment starting at “6:40” becomes “6:50”). Malformed JSONs (11.7\%) were wrapped in a ``raw\_output'' field. For gaze prediction, which involves anticipating future events, only the preceding 400s were used. For {KnowledgeNet graph generation} (see Section~\ref{sec:knowledgenet}), Gemini was prompted to identify relevant kitchen objects from bounding boxes, timestamps, or visual context. The model returned a plain Python list of object names, guided by question-specific prompts. 
Gemini Flash 2.0 supports a maximum input duration of ~45 minutes. To meet this, we applied a temporal divisor that adaptively accelerated videos (sampled at 1 FPS), with a minimum duration of one second. All videos were standardized to a 2400-second (40-minute) processing window to ensure compatibility.

\paragraph{Configuration Details. }
During {inference}, for any given question, we use one of these input configurations:
\begin{itemize}
    \item \textbf{Video Only}: The model receives the natural language question and the raw video, without any structured graph representations. If the question input specified a particular clip or image, that corresponding video segment or frame was provided. 
    \item \textbf{SceneNet (S-Net)}: The model receives the question along with the full set of (globally timestamp-normalized) scene graphs corresponding to all segments of the referenced video. Raw video was not used in this configuration.
    \item \textbf{KnowledgeNet (K-Net)}: The model receives as input the question, the video and the set of textualized semantic paths from K-Net.
\end{itemize}
For both SceneNet and KnowledgeNet configurations, if the input explicitly specified an image, or if the question included a single \texttt{<TIME>} tag, that corresponding frame was extracted from the video. If a bounding box tag (\texttt{<BBOX>}) was also included in the question, this bounding box was drawn onto the extracted image. This image was then provided to the MLLM alongside the respective graph representation (S-Net or K-Net) and the question to aid in visual disambiguation or grounding.

\subsection{Results}
\label{sec:results}


\textbf{Per-category results.} Table~\ref{tab:category_results} presents per-category results for the Video Only, SceneNet, and KnowledgeNet models. 
{S-Net} achieves a {2.9\% improvement} in overall performance w.r.t. Video Only, with 12\% improvement in 3D perception and Object Motion categories and 22\% improvement on Recipe. These results suggest that scene graphs help the model better capture spatio-temporal dynamics especially useful for modeling objects changing location through time. SceneNet performs well across most categories, though slightly underperforms in \textit{action}, \textit{gaze}, and \textit{ingredient}, where scene graphs may offer limited value due to the need for fine-grained visual or compositional cues. {K-Net} improves performance over Video Only by {3\% overall}, with consistent gains across all categories except Action. 
Notable gains appear in Recipe and Nutrition categories, which benefit from external procedural knowledge beyond visual detail.


\begin{table}[t]
\small
    \centering
    \resizebox{.95\linewidth}{!}{%
    \begin{tabular}{@{}lccc@{}}
        \toprule
        \textbf{Category} & \textbf{Video Only} & \textbf{SceneNet} & \textbf{KnowledgeNet} \\
        \midrule
        3D Perception & 29.38 & \textbf{41.84} \textcolor{green!60!black}{\small$\uparrow$12.46} & 34.29 \textcolor{green!60!black}{\small$\uparrow$4.91} \\
        Action & \textbf{48.05} & 31.97 \textcolor{red}{\small$\downarrow$16.08} & 47.86 \textcolor{red}{\small$\downarrow$0.19} \\
        Gaze & 30.45 & 23.75 \textcolor{red}{\small$\downarrow$\textcolor{white}{x}6.70} & \textbf{31.45} \textcolor{green!60!black}{\small$\uparrow$1.00} \\
        Ingredient & 45.17 & 39.67 \textcolor{red}{\small$\downarrow$\textcolor{white}{x}5.50} & \textbf{45.83} \textcolor{green!60!black}{\small$\uparrow$0.66} \\
        Nutrition & 33.67 & 34.67 \textcolor{green!60!black}{\small$\uparrow$\textcolor{white}{x}1.00} & \textbf{38.00} \textcolor{green!60!black}{\small$\uparrow$4.33} \\
        Object Motion & 17.53 & \textbf{30.17} \textcolor{green!60!black}{\small$\uparrow$12.64} & 19.36 \textcolor{green!60!black}{\small$\uparrow$1.83} \\
        Recipe & 40.75 & \textbf{63.63} \textcolor{green!60!black}{\small$\uparrow$22.88} & 49.50 \textcolor{green!60!black}{\small$\uparrow$8.75} \\
        \midrule
        \textbf{Overall} & 35.00 & 37.96 \textcolor{green!60!black}{\small$\uparrow$2.96} & \textbf{38.04} \textcolor{green!60!black}{\small$\uparrow$3.04} \\
        \bottomrule
    \end{tabular}
    }
    \caption{Per-category accuracy (\%) across configurations. Green $\uparrow$ indicates improvement over the Video Only baseline; red $\downarrow$ indicates a decrease.}
    \label{tab:category_results}
\end{table}

\begin{figure}[t]
    \centering
    \includegraphics[width=\linewidth]{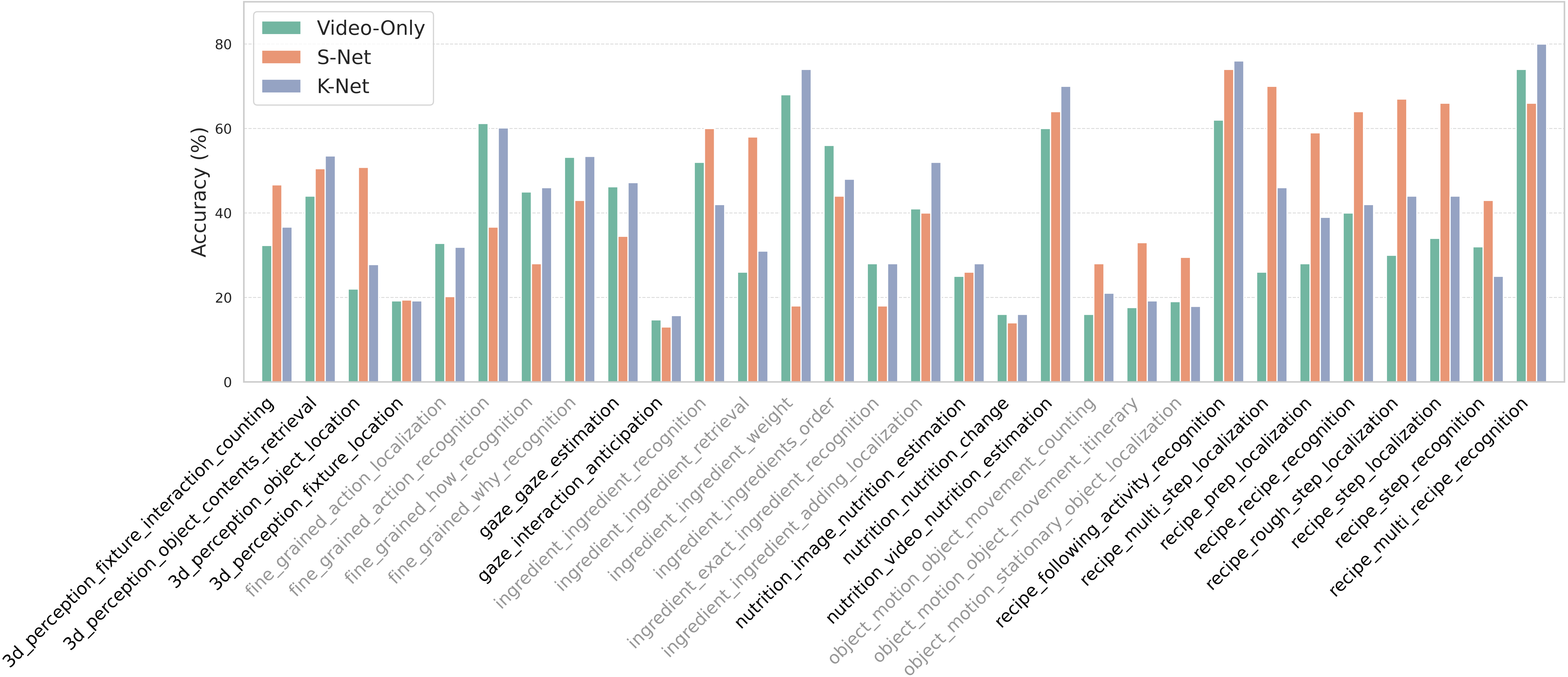}    \caption{Accuracy (\%) for each micro-category. }
    \label{fig:microcategory_accuracies}
\end{figure}

\begin{figure}[h]
    \centering
    \includegraphics[width=\linewidth]{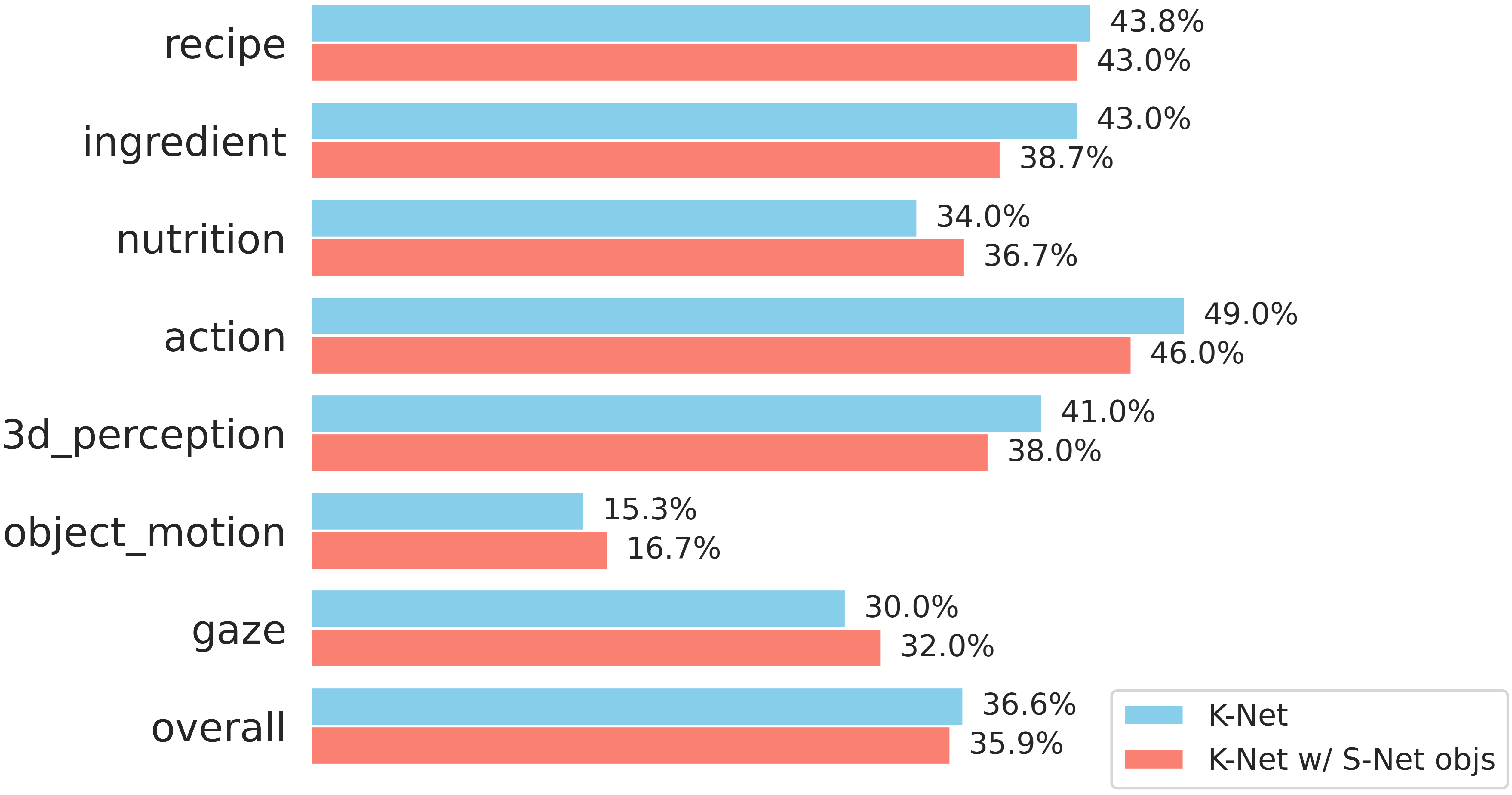}
    \caption{Per-category accuracy (\%) for K-Net w/ video and K-Net w/ video + SceneNet objects.}
    \label{fig:barplot_k-net}
\end{figure}

\noindent \textbf{Micro-category Analysis. }Figure~\ref{fig:microcategory_accuracies} shows the performance of all configurations across all micro-categories within the broader category groups. The results highlight that different approaches excel in different contexts. In particular, some methods consistently outperform others across all micro-categories within specific macro-categories. For example, S-Net shows consistent improvements across all micro-categories of 3D Perception, while K-Net achieves similar consistent gains in the Recipe category.

\subsection{Ablations}

\begin{table}[t]
    \centering
    \small
    \begin{tabular}{@{}l|cc|cc|cc@{}}
        \toprule
        \multicolumn{1}{l|}{\textbf{Model}} 
        & \multicolumn{2}{c|}{\textbf{S-Net}} 
        & \multicolumn{2}{c|}{\textbf{K-Net}} 
        & \multicolumn{2}{c}{\textbf{S+K-Net}} \\
        \cmidrule(lr){1-1} \cmidrule(lr){2-3} \cmidrule(lr){4-5} \cmidrule(lr){6-7}
        \textbf{Video} 
        & \textbf{\ding{51}} & \textbf{\ding{55}} 
        & \textbf{\ding{51}} & \textbf{\ding{55}} 
        & \textbf{\ding{51}} & \textbf{\ding{55}} \\
        \midrule
        3D Perception & 34.50 & \textbf{42.50} & \textbf{41.00} & 31.00 & \textbf{37.50}& 36.50 \\
        Action        & \textbf{40.50} & 33.00 & \textbf{49.00} & 22.00 & \textbf{43.50}& 25.50 \\
        Gaze          & 27.00 & \textbf{30.00} & \textbf{30.00} & 23.00 & \textbf{25.00}& 21.00 \\
        Ingredient    & \textbf{40.66} & 38.67 & \textbf{43.00} & 24.00 & \textbf{41.00}& 22.67 \\
        Nutrition     & 30.66 & \textbf{36.00} & \textbf{34.00}& 28.67 & \textbf{34.67}& 30.67 \\
        Object Motion & 26.00 & \textbf{29.33} & 15.33 & \textbf{31.33}& 17.33 & \textbf{31.33}\\
        Recipe        & 49.75 & \textbf{62.75} & \textbf{43.75}& 30.00 & \textbf{46.75}& 33.25 \\
        \midrule
        \textbf{Overall} & 35.58 & \textbf{38.89} & \textbf{36.58}& 27.14 & \textbf{35.11}& 28.70 \\
        \bottomrule
    \end{tabular}%
    \caption{Per-category accuracy (\%) for S-Net, K-Net, and their combination (S+K-Net).} 
    \label{tab:ablation}
\end{table}



We conducted ablation studies using 50 samples per micro-category. Detailed results are presented in Table \ref{tab:ablation}.

\textbf{S-Net (w/ and w/o video).}
We first evaluated the impact of adding raw video to S-Net. Overall, S-Net without video outperformed the combined S-Net with video setup, suggesting that S-Net’s structured information is often sufficient, while raw video can introduce noise or complicate fusion. However, for inherently visual categories like Action and Ingredient, S-Net with video showed clear gains, consistent with the strengths of the video-only baseline. Despite these cases, the general trend favored S-Net only.

\textbf{K-Net (w/ and w/o video).}
We then compared K-Net with and without raw video. The version without video generally underperformed its video-augmented counterpart, highlighting the importance of grounding symbolic knowledge in visual context to avoid reasoning using just commonsense knowledge not grounded in the video.

\textbf{S+K-Net.}
This joint setup generally underperformed compared to using either source alone, likely due to the inclusion of excessive or irrelevant information that hinders reasoning. It also reflects the individual weaknesses of its components, i.e., SceneNet's sensitivity to added video and KnowledgeNet's reliance on visual grounding. These results highlight the challenges of integrating heterogeneous knowledge in VQA. We hypothesize that a more selective, question-aware fusion strategy could improve alignment with task-specific needs and enhance performance.

\textbf{K-Net (from S-Net objects).}
We experimented with generating K-Net graphs using objects derived from S-Net entities, aiming for tighter integration through contextually grounded ConceptNet graphs (Figure \ref{fig:barplot_k-net}). However, this setup underperformed compared to more targeted KnowledgeNet approaches, likely due to the added noise from including visually grounded but question-irrelevant entities.

\subsection{Submission Strategy}

For the final challenge submission, corresponding to team name \textit{DeepFrames}, we selected the best-performing method for each micro-category. This ensemble strategy allowed each question to benefit from the input modality best suited to its specific reasoning needs.

\section{Conclusion}
\label{sec:conclusion}
In this report, we present our submission to the HD-EPIC VQA challenge. We propose extracting graph-based structured representations from video using two modules: SceneNet and KnowledgeNet. Each offers complementary strengths for egocentric video question answering, and we alternate between them based on the question category. A promising direction for future work is to integrate both modules into a unified approach, leveraging their strengths jointly rather than independently.


\noindent\textbf{Acknowledgements.} This work was conducted at the Smart Eyewear Lab, a joint research center between EssilorLuxottica and Politecnico di Milano.

{
    \small
    \bibliographystyle{ieeenat_fullname}
    \bibliography{main}
}


\end{document}